\documentclass{article}
\pdfoutput=1

\usepackage[preprint,nonatbib]{nips_2018}

\usepackage[utf8]{inputenc} % allow utf-8 input
\usepackage[T1]{fontenc}    % use 8-bit T1 fonts
\usepackage{hyperref}       % hyperlinks
\usepackage{url}            % simple URL typesetting
\usepackage{booktabs}       % professional-quality tables
\usepackage{amsfonts}       % blackboard math symbols
\usepackage{nicefrac}       % compact symbols for 1/2, etc.
\usepackage{microtype}      % microtypography

\usepackage{graphicx}
\usepackage{bm}
\usepackage{amsmath}
\usepackage{mathtools}  
\usepackage{xcolor}
\usepackage{bm}
\usepackage{bbm}
\usepackage{amssymb}
\usepackage{amsthm}

\usepackage[noline, boxed]{algorithm2e}

\title{Model-Based Regularization for Deep Reinforcement Learning with Transcoder Networks}

\author{
  Felix Leibfried, Peter Vrancx \\
  PROWLER.io\\
  Cambridge, UK\\
  \texttt{\{felix,peter\}@prowler.io} \\
}

\begin{document}

\maketitle

\begin{abstract}
  This paper proposes a new optimization objective for value-based deep reinforcement learning. We extend conventional Deep Q-Networks (DQNs) by adding a model-learning component yielding a transcoder network. The prediction errors for the model are included in the basic DQN loss as additional regularizers. This augmented objective leads to a richer training signal that provides feedback at every time step. Moreover, because learning an environment model shares a common structure with the RL problem, we hypothesize that the resulting objective improves both sample efficiency and performance. We empirically confirm our hypothesis on a range of 20 games from the Atari benchmark attaining superior results over vanilla DQN without model-based regularization.
\end{abstract}

\section{Introduction}
\label{introduction}

Reinforcement learning (RL) is an area of machine learning that enables artificial agents to identify optimal behavioral policies through interactions with their environment \cite{Sutton1998}. RL has been successfully applied in high-dimensional problems, such as the Arcade Learning Environment (ALE) for Atari games \cite{Bellemare2013}. Recent methods combining RL with deep learning have demonstrated super-human performance \cite{Mnih2015} in these settings. However, when the reward signal is sparse, RL is often sample-inefficient. When the reward is absent most of time, there is no training feedback to reinforce the agent's behavior, resulting in slow learning and sub-optimal performance.

In this paper, we aim to address the problem of sparse reward signals by combining model-free value learning with predictive model learning that provides a richer training signal. We propose a hybrid model and value learning objective that both improves performance and reduces sample complexity. Combining model and value learning can be motivated by examining the relationship between both objectives. Previous research has demonstrated that model learning shares an inherent structure with the RL problem \cite{Parr2008,Sutton2012,Song2016}. Song et al. \cite{Song2016} show that learning features for linear value function approximation by minimizing the model prediction error directly leads to a lower Bellman error. Unfortunately, these results apply to linear approximation and do not generalize to the deep non-linear architectures we consider here \cite{Song2016}.  Therefore, we propose an alternative approach: the incorporation of model predictions into the value learning process.  Rather than defining a separate feature learning pipeline based on the model prediction error, we combine both learning problems by using the model prediction error as a regularizer for the value learning objective. Including this model-based regularization yields a transcoder network that provides feedback at every time step based on the current prediction error. This richer feedback signal eliminates one source of sample inefficiency caused by sparse rewards.

We successfully combine a predictive environment model with a reward-maximizing agent in order to improve sample efficiency and performance on a wide range of Atari games. To this end, we phrase a novel optimization objective for deep RL that includes augmentations to learn an environment model, in addition to an optimal policy. We design a parametric agent that shares parameters for optimal acting with a model for environment prediction. In ALE experiments, our approach outperforms RL without model-based regularization by large margins. In short, our contributions are as follows:
\begin{itemize}
\item propose a novel joint optimization objective for deep RL and environment prediction;
\item propose a novel deep transcoder architecture sharing parameters for acting and prediction;
\item demonstrate accurate future video frame and reward prediction in ALE;
\item and outperform deep RL without model-based regularization on several Atari games.
\end{itemize}

\section{Background}
\label{background}

We consider a discrete-time, infinite-horizon, discounted Markov Decision Process (MDP) $(S, A, P^{(R)}, P^{(S)}, \gamma)$.
Here $S$ denotes the state space, $A$ the set of possible actions,  $P^{(R)}:S \times A \times R \rightarrow [0,1]$ is the reward function, $P^{(S)}: S \times A \times S \rightarrow [0,1]$ is the state transition function, and $\gamma \in (0,1)$ is a discount factor. An agent, being in state $\bm{S}_t \in S$ at time step $t$, executes an action $\bm{a}_t \in A$ sampled from its policy $\pi(\bm{a}_t|\bm{S}_t)$, a conditional probability distribution  where $\pi:S \times A \rightarrow [0,1]$. The agent's action elicits from the environment a reward signal $r_t \in R$ indicating instantaneous reward, a terminal flag $f_{t} \in \{0,1\}$ indicating a terminal event that restarts the environment, and a transition to a successor state $\bm{S}_{t+1} \in S$. We assume that the sets $S$, $A$, and $R$ are discrete. The reward $r_t$ is sampled from the conditional probability distribution $P^{(R)}(r_t|\bm{S}_t,\bm{a}_t)$. Similarly, $f_{t} \sim P^{(F)}(f_{t} | \bm{S}_t,\bm{a}_t)$ with $P^{(F)}: S \times A \times \{0,1\} \rightarrow [0,1]$, where a terminal event ($f_t = 1$) restarts the environment according to some initial state distribution $\bm{S}_0 \sim P^{(S)}_0(\bm{S}_0)$. The state transition to a successor state is determined by a stochastic state transition function according to $\bm{S}_{t+1} \sim P^{(S)}(\bm{S}_{t+1} | \bm{S}_t,\bm{a}_t)$. 

The agent's ultimate goal is to maximise future cumulative reward 
\[ \mathbb{E}_{P^{(S)}_0,\pi,P^{(R)},P^{(F)},P^{(S)}} \left [ \sum_{t=0}^{\infty} \left ( \prod_{t'=0}^{t} (1-f_{t'-1}) \right ) \gamma^t r_t \right ] \]
with respect to the policy $\pi$. An important quantity in RL are Q-values $Q^{(\pi)}(\bm{S}_t,\bm{a}_t)$, which are defined as the expected future cumulative reward when executing action $\bm{a}_t$ in state $\bm{S}_t$ and subsequently following policy $\pi$. Q-values enable us to conveniently phrase the RL problem as $\max_{\pi} \sum_{\bm{a}} \pi(\bm{a}|\bm{S}) Q^{(\pi)}(\bm{S},\bm{a})$ for all $\bm{S}$.

Value-based reinforcement learning approaches identify optimal Q-values directly using parametric function approximators $Q_{\bm{\theta}}(\bm{S},\bm{a})$, where $\bm{\theta} \in \bm{\Theta}$ represents the parameters  \cite{Watkins1989,Busoniu2010}. Optimal Q-value estimates then correspond to an optimal policy $\pi(\bm{a}|\bm{S}) = \delta_{\bm{a}, \text{argmax}_{\bm{a}'} Q_{\bm{\theta}}(\bm{S},\bm{a}')}$. Deep Q-networks \cite{Mnih2015} learn a deep neural network based  Q-value approximation by performing stochastic gradient descent on the following training objective:  

\begin{equation}
\label{eq:q-loss}
L^{(Q)}(\bm{\theta})  = \mathbb{E}_{\bm{S}, \bm{a}, r, f, \bm{S}'} \bigg[ \bigg(r +  (1-f) \gamma \max_{\bm{a}'} Q_{\bm{\theta}^-}(\bm{S}',\bm{a}')
  - Q_{\bm{\theta}}(\bm{S},\bm{a}) \bigg)^2 \bigg] ,
\end{equation}

The expectation ranges over transition samples $\bm{S}, \bm{a}, r, f, \bm{S}'$ sampled from an experience replay memory ($\bm{S}'$ denotes the state at the next time step). Use of this replay memory, together with the use of a separate target network $Q_{\bm{\theta}^-}$ (with different parameters $\bm{\theta}^- \in \bm{\Theta}$) for calculating the bootstrap values $ \max_{\bm{a}'} Q_{\bm{\theta}^-}(\bm{S}',\bm{a}')$, helps stabilize the learning process.

\section{Related Work}
\label{model-based_rl}

While model-free RL aims to learn a policy directly from transition samples, model-based RL attempts to speed up the learning process by learning about both the environment and the actual RL task. In robotics, there is a range of model-based RL approaches that have been successfully deployed \cite{Deisenroth2011,Levine2013,Levine2014,Heess2015,Pong2018}, even for visual state spaces \cite{Wahlstrom2015,Watter2015,Finn2016,Levine2016}. 
However, in other vision-based domains (e.g. Atari), model-based RL has been less successful, despite plenty of model-learning approaches that demonstrably learn accurate environment models \cite{Oh2015,Fu2016,Chiappa2017,Leibfried2017,Wang2017,Weber2017,Buesing2018}.

There are several ways environment models can be used in RL. Planning approaches use the environment model to solve a virtual RL problem and act accordingly in the real environment \cite{Wang2009,Browne2012,Russell2016}. DYNA-style learning \cite{Sutton1990,Silver2008,Sutton2012} augments the dataset with virtual samples for RL training. These virtual samples are generated from a learned environment model and are combined with samples from the actual environment to produce the final learning update. Explicit exploration approaches use models to direct exploration. They encourage the agent to take actions that most likely lead to novel environment states \cite{Stadie2015,Oh2015,Pathak2017}. 

Here, we focus on another approach that trains a predictive model to extract features for value function learning.  Recent research has identified a relation between model and value function approximation \cite{Parr2008, Song2016}. This work provides a theoretical basis for feature learning that connects the model prediction error to the Bellman error for linear value function approximation. Unfortunately, these results do not carry over to non-linear architectures. We provide an alternative approach for joint model and value function learning, and show that this improves performance over vanilla deep RL without model-based regularization. \textit{Note that this is in general not considered as model-based RL since the predictive model is merely used for feature extraction but not for planning.}

Closely related to our work is the UNREAL agent in Jaderberg et al. \cite{Jaderberg2017} where model-free RL is combined with auxiliary losses to improve performance of deep RL in an actor-critic setting. Also closely related is the approach from Alaniz \cite{Alaniz2017} attaining impressive results by combining model-based learning with planning in the visual domain of Minecraft. The work of Oh et al. \cite{Oh2017} is similar, too, extending model-free RL with an abstract model operating in latent space to predict rewards and values, but not actual next environmental states.

\section{Model-Based Regularization}
\label{model_based_regularization}

The problem of sparse rewards can be addressed by providing additional training signals in the course of the learning process that occur at a higher frequency. One way to provide such training signals is via a shared parametric model for both reward maximization and environment prediction. The idea is that the problem of environment prediction has a common latent structure with the actual RL problem leading to an internal representation that is useful for reward maximization \cite{Parr2008,Song2016}. This can be enabled by extending DQNs\footnote{We present our approach using the vanilla DQN framework as originally proposed by \protect{\cite{Mnih2015}}. The current state-of-the-art framework for value-based deep RL is the Rainbow framework \protect{\cite{Hessel2018}}, which is a combination of several independent DQN improvements. However, because these extensions do not use an environment model, the approach presented in this paper is orthogonal to the individual Rainbow components and could be combined with Rainbow.} to predict future environmental states in addition to Q-values. In the following, we propose a novel deep transcoder architecture (Section~\ref{network_architecture}) and optimization objective (Section~\ref{optimization_objective}) for jointly learning environment dynamics and Q-values with the aim to improve both sample efficiency and performance in deep RL. Our algorithm for training the transcoder agent is presented in Section~\ref{algorithm}

\subsection{Network Architecture}
\label{network_architecture}

\begin{figure}[t]
\begin{center}
\centerline{\includegraphics[width=.9\columnwidth]{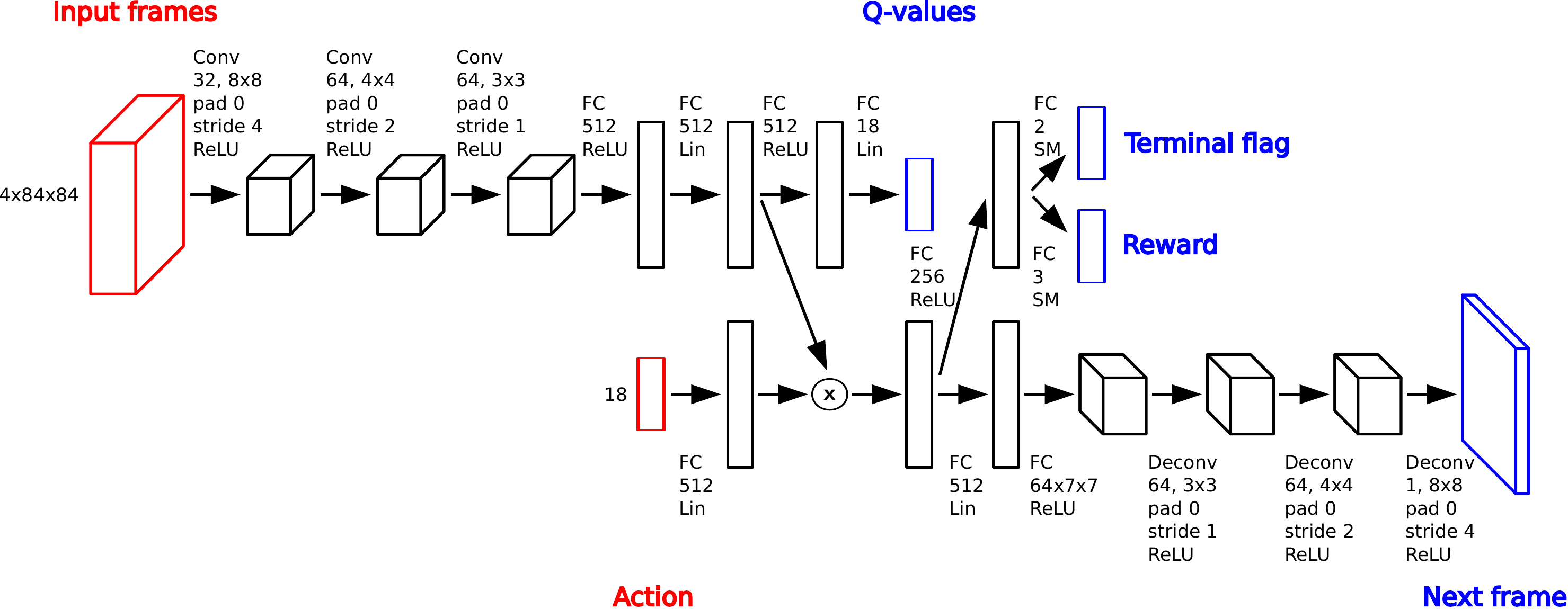}}
\caption{Network architecture. The base network is a deep Q-network that maps the current state given by the last four video frames to Q-value estimates for each potential action. This base network is extended by an action-conditioned path that enables the estimation of the current terminal flag, the instantaneous reward and the next-time-step video frame. Network inputs are colored in red and network outputs in blue. Computational units comprise convolutional ('Conv'), ordinary forward ('FC'), and deconvolutional ('Deconv') connections that are combined with linear ('Lin'), rectified linear ('ReLU'), or softmax ('SM') activations. The operator $\times$ denotes element-wise multiplication.}
\label{fig:architecture}
\end{center}
\end{figure}

Our proposed network architecture is depicted in Figure~\ref{fig:architecture} and comprises two components. The first component is action-unconditioned and maps the state of the environment to Q-value estimates for each potential action the agent \textit{could take}. The second component is action-conditioned and uses, in addition to the state of the environment, the action \textit{actually taken} by the agent in order to make predictions about the state at the next time step as well as the reward and the terminal flag.

On a more detailed level, there are five information-processing stages. The first stage is an encoding that maps the state of the environment at time step $t$, a three-dimensional tensor $\bm{S}_{t}$ comprising the last $h$ video frames to an internal compressed feature vector $\bm{h}_{t}^{\text{enc}}$ via a sequence of convolutional and ordinary forward connections. The second stage is the Q-value prediction that maps the internal feature vector $\bm{h}_{t}^{\text{enc}}$ to Q-value predictions $Q(\bm{S},\bm{a})$ for each possible action $\bm{a}$ that the agent could potentially take. The Q-value prediction path consists of two ordinary forward connections. The third stage transforms the hidden encoding $\bm{h}_{t}^{\text{enc}}$ into an action-conditioned decoding $\bm{h}_{t}^{\text{dec}}$ by integrating the action $\bm{a}_t$ actually taken by the agent. This process first transforms the action into a one-hot encoding followed by a forward connection and an element-wise vector multiplication with $\bm{h}_{t}^{\text{enc}}$ leading to $\bm{h}_{t}^{\text{dec}}$. Note that the two layers involved in this element-wise vector multiplication are the only two layers in the network without bias neurons. The fourth stage maps the action-conditioned decoding $\bm{h}_{t}^{\text{dec}}$ into a prediction for the terminal flag $f_t$ and the instantaneous reward $r_t$  in a sequence of forward connections. Both the terminal flag $f_t$ and the reward $r_t$ are categorical variables. The terminal flag is binary and the reward is ternary because reward values from ALE are clipped to lie in the range $\{-1,0,+1\}$ \cite{Mnih2015}. The last stage maps $\bm{h}_{t}^{\text{dec}}$ to the video frame $\bm{S}^{'}_{t+1}[-1]$ at the next time step $t+1$ by using forward and deconvolutional connections \cite{Dosovitskiy2015}.

The network uses linear, rectified linear \cite{Glorot2011} and softmax activities. The video frames fed into the network are $84 \times 84$ grayscale images (pixels $\in [0,1]$) down-sampled from the full $210 \times 160$ RGB images from ALE. Following standard literature \cite{Mnih2015,Oh2015}, the video frame horizon is $h=4$ and frame skipping is applied. Frame skipping means that at each time step when executing an action in the environment, this action is repeated four times and frames with repeated actions are skipped. The instantaneous rewards are accumulated over skipped frames.

\subsection{Optimization Objective}
\label{optimization_objective}

The optimization objective for training the network comprises four individual loss functions that are jointly minimized with respect to the network weights $\bm{\theta}$: one for Q-value prediction $L^{(Q)}(\bm{\theta})$ and three additional loss functions. The first additional loss function is for predicting the terminal flag $L^{(F)}(\bm{\theta})$, the second for predicting the instantaneous reward $L^{(R)}(\bm{\theta})$, and the third for predicting the video frame at the next time step $L^{(S)}(\bm{\theta})$. All these parts are additively connected leading to a compound loss
\begin{equation}
\label{eq:compound_loss}
L(\bm{\theta}) = L^{(Q)}(\bm{\theta}) + \lambda^{(F)} L^{(F)}(\bm{\theta}) + \lambda^{(R)} L^{(R)}(\bm{\theta}) + \lambda^{(S)} L^{(S)}(\bm{\theta}),
\end{equation}
where $\lambda^{(F)}$, $\lambda^{(R)}$, and $\lambda^{(S)}$ are positive coefficients to balance the individual parts. The compound loss is an off-policy objective that can be trained with a set of environment interactions obtained from any policy. The individual parts of Equation~\eqref{eq:compound_loss} can then be expressed as expected values over transitions  of the form $\bm{S}, \bm{a}, r, f, \bm{S}'$ that are sampled from a replay memory.

The Q-value prediction loss can be written as in Equation~\eqref{eq:q-loss}, whereas the terminal flag prediction loss is given by
\begin{equation}
\label{eq:terminal_flag_loss}
L^{(F)}(\bm{\theta}) = \mathbb{E}_{\bm{S}, \bm{a}, f} \Bigg[ -\ln P^{(F)}_{\bm{\theta}}(f|\bm{S},\bm{a}) \Bigg],
\end{equation}
where $P^{(F)}_{\bm{\theta}}$ refers to the terminal flag predictor in Figure~\ref{fig:architecture} for predicting the terminal flag given a specific state and action. Note that the terminal flag is a binary quantity and $P^{(F)}_{\bm{\theta}}$ is a parametric categorical probability distribution.

Similarly, the loss for instantaneous reward prediction can be expressed as
\begin{equation}
\label{eq:reward_loss}
L^{(R)}(\bm{\theta}) = \mathbb{E}_{\bm{S}, \bm{a}, r} \Bigg[ -\ln  P^{(R)}_{\bm{\theta}}(r|\bm{S},\bm{a}) \Bigg],
\end{equation}
where $P^{(R)}_{\bm{\theta}}$ is a state-action-conditioned parametric categorical distribution over the reward signal.

The loss for predicting the video frame at the next time step can be formulated as
\begin{equation}
\label{eq:video_frame_loss}
L^{(S)}(\bm{\theta}) = \mathbb{E}_{\bm{S}, \bm{a}, f, \bm{S}'} \Bigg[  \frac{1}{2} (1 - f)\big|\big| M^{(S)}_{\bm{\theta}}(\bm{S},\bm{a}) - \bm{S}'[-1] \big|\big|_\mathsf{F}^2 \Bigg],
\end{equation}
where $M^{(S)}_{\bm{\theta}}$ refers to a deterministic parametric map to predict the next state observation
(denoted $\bm{S}'[-1]$), given the current state and the action (i.e. predict the next video frame given the last four video frames and action in Atari). $||\cdot||_\mathsf{F}^2$ refers to the squared Frobenius norm between two images. 

\subsection{Algorithm}
\label{algorithm}

Our algorithm commences in rounds as depicted in Algorithm~\ref{pseudocode}. At each time step, the agent observes the current environmental state and takes an epsilon-greedy action with respect to the current Q-values. This leads the environment to elicit a reward signal, a terminal flag, and transition to a successor state. An experience replay memory is updated accordingly. Every fourth interaction with the environment, the agent samples experiences from the experience replay to perform a gradient update step on the loss from Equation~\eqref{eq:compound_loss}. The target network is periodically updated.

\begin{algorithm}
\caption{Pseudocode for our algorithm.}
\label{pseudocode}
initialize replay memory $\mathbb{M}$ and network weights $\bm{\theta}$\\
initialize target network weights $\bm{\theta}^{-} = \bm{\theta}$  and $t=0$ \\
start environment, yielding an initial $\bm{S}$ \\
\While{$t<50,000,000$}{
observe $\bm{S}$ and
choose $\bm{a}$ epsilon-greedy from $Q_{\bm{\theta}}(\bm{S},\bm{a})$\\
execute $\bm{a}$ and observe $r$, $f$ and $\bm{S}'$ \\
add $\left\langle\bm{S}, \bm{a}, r, f, \bm{S}'\right\rangle$-tuple to $\mathbb{M}$\\
\If{$t\%4==0$}{
   sample $\left\langle\bm{S}^{(i)}, \bm{a}^{(i)}, r^{(i)}, f^{(i)}, \bm{S}'^{(i)}\right\rangle_{i=1}^{I}$ from $\mathbb{M}$\\
perform gradient update step on $L(\bm{\theta})$ \\
   }
set $\bm{\theta}^{-} = \bm{\theta}$ \ \textbf{if} \ $t\%32,000 == 0$\\
set $t = t+1$\\
set $\bm{S} = \bm{S}'$ \ \textbf{if} \ $f==0$ \ \textbf{else} \ restart environment}
\end{algorithm}

\section{Experiments}
\label{experiments}

In the following, we give a brief summary of training details (Section~\ref{training}) and subsequently present our empirical results. The empirical results comprise the visualization of model predictions (Section~\ref{model_predictions}), a detailed analysis of the individual components of the loss function (Section~\ref{loss_components}), a quantification of game play performance (Section~\ref{game_play performance}) as well as demonstrating sample efficiency (Section~\ref{sample_efficiency}).

\subsection{Training Details}
\label{training}

The agent follows an epsilon-greedy policy with linear epsilon-annealing over one million steps from $1.0$ to $0.1$. Agent parameters are represented by a deep neural network as in Figure~\ref{fig:architecture} with randomly initialized weights according to \cite{Glorot2010,Oh2015}. The network is trained for $5 \cdot 10^7$ time steps. The target network is updated every $32,000$ steps \cite{Hessel2018}. Training and epsilon-annealing start after $50,000$ steps. When there is a terminal event in the environment, or the agent loses one life, or an episode exceeds a length of $27,000$ steps, the environment is randomly restarted \cite{Hessel2018}. Random restarting means randomly sampling up to $30$ NOOP-actions at the beginning of the episode. Environment interactions are stored in a replay memory comprising at most the last $10^6$ time steps.

Network parameters are updated every fourth environment step by sampling a mini-batch of $32$ trajectories with length $4$ from the replay memory---note that in Section~\ref{optimization_objective}, we omit the temporal dimension to preserve a clearer view. In practical terms, trajectory samples can enable a more robust objective for learning Q-values by creating multiple different temporally extended target values for one prediction \cite{Sutton1998,Hessel2018}.
Mini-batch samples are used to optimize Equation~\eqref{eq:compound_loss} ($\lambda^{(F)}=\lambda^{(R)}=1, \lambda^{(S)}=\frac{1}{84}$) by taking a single gradient step with Adam \cite{Kingma2015} (learning rate $6.25\cdot 10^{-5}$, gradient momentum 0.95, squared gradient momentum 0.999 and epsilon $1.5\cdot 10^{-4}$). Gradients are clipped when exceeding a norm of $1.0$ \cite{Leibfried2017}. In practice, the Q-value prediction loss as well as the loss for terminal flag and reward prediction is clipped for large errors (exceeding $1.0$ \cite{Mnih2015}) and small probability values (below $\text{e}^{-10}$ \cite{Leibfried2017}) respectively. Because non-zero rewards and terminal events occur less frequently than zero rewards and non-terminal events, they are weighted inversely proportionally to their relative occurrence.

We empirically validate our approach in the Atari domain \cite{Bellemare2013}. We compare against ordinary deep RL without any environment model (DQN, \cite{Mnih2015}). Our aim is to verify that the model-regularized objective improves learning results over basic DQN. We show that our approach leads to significantly better game play performance and sample efficiency across 20 Atari games.

\begin{figure}[t]
\begin{center}
\centerline{\includegraphics[width=1\columnwidth]{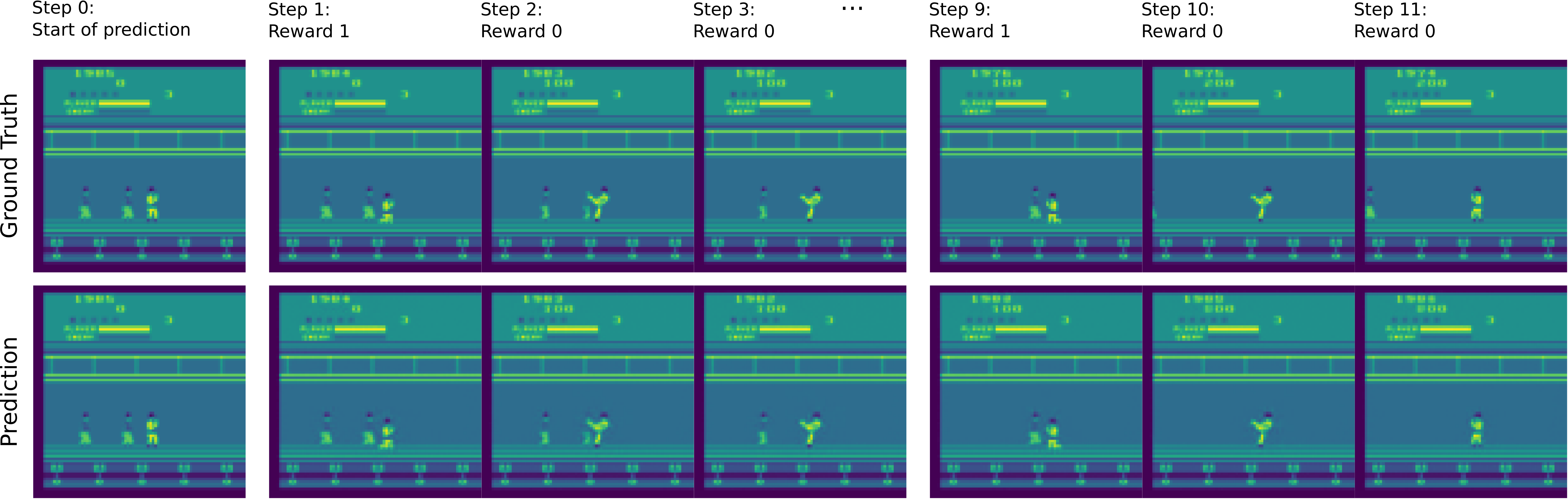}}
\caption{Predictions in Kung Fu Master. The top row shows the ground truth over eleven time steps (44 frames) and the bottom row shows the action-conditioned predictions of our approach. Predictions remain accurate when unrolling the network over time. \textit{In this example, the reward signal is perfectly predicted}. In time steps 10 and 11, a random opponent enters the scene from the left which is not predictable when merely utilizing information from the start of the prediction at step 0. }
\label{fig:predictions_kfm}
\end{center}
\end{figure}

\subsection{Visualized Model Predictions}
\label{model_predictions}

As a proof of concept, in Figure~\ref{fig:predictions_kfm}, we visualize model predictions on the game Kung Fu Master and compare to ground truth video frames over a time horizon of eleven steps (44 frames) starting from a given initial state (after executing the policy for $100$ steps under the random NOOP condition). When virtually unrolling the network, predicted frames are clipped to $[0, 1]$ \cite{Leibfried2017}. \textit{In the example depicted, the reward signal is perfectly predicted}. Model-predicted video frames accurately match ground truth frames. At time steps 10 and 11, a random object is entering the scene from the left, which our model cannot predict because there is no information available at time step 0 to foresee this event. This in accordance with previous findings \cite{Oh2015,Leibfried2017}.

\subsection{Components of the Loss Function}
\label{loss_components}

\begin{figure}[t]
\begin{center}
\centerline{\includegraphics[width=1\columnwidth]{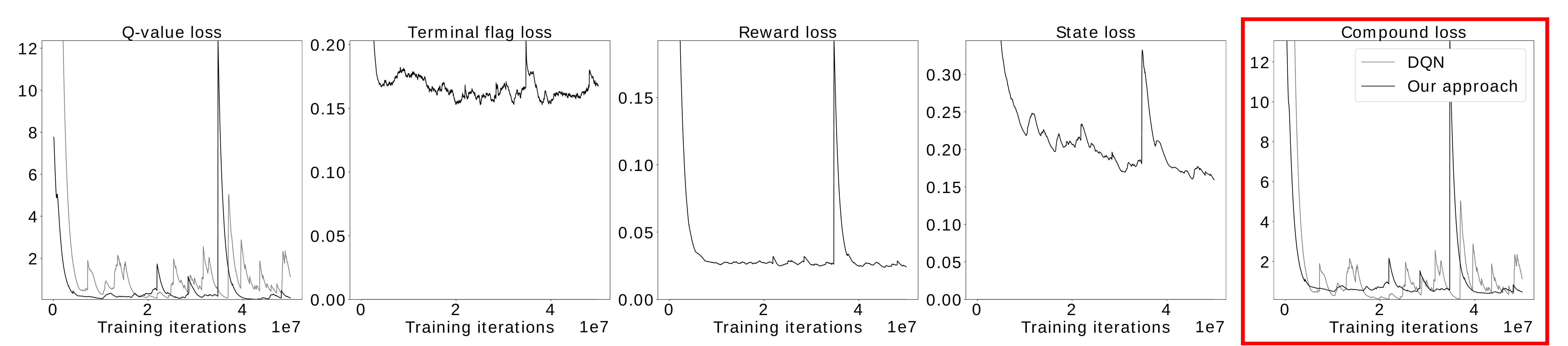}}
\caption{Training loss in Kung Fu Master. All individual loss parts from Equation~\eqref{eq:compound_loss} are depicted as a function of training iterations. The loss of the DQN baseline (gray) consists only of the Q-value loss from Equation~\eqref{eq:q-loss}. The compound loss of our approach (black) is dominated by the Q-value loss and only mildly affected by the augmented loss parts that act as a regularizer.}
\label{fig:training_loss_kfm}
\end{center}
\end{figure}

We analyze the different components of the loss proposed in Equation~\eqref{eq:compound_loss}. As an illustrative example, we report the losses during training on the game Kung Fu Master (see Figure~\ref{fig:training_loss_kfm}). Clearly, the compound loss is dominated by the Q-value loss and all other loss parts act as a regularizer.

\subsection{Game Play Performance}
\label{game_play performance}

\begin{figure}[h!]
\begin{center}
\centerline{\includegraphics[width=1\columnwidth]{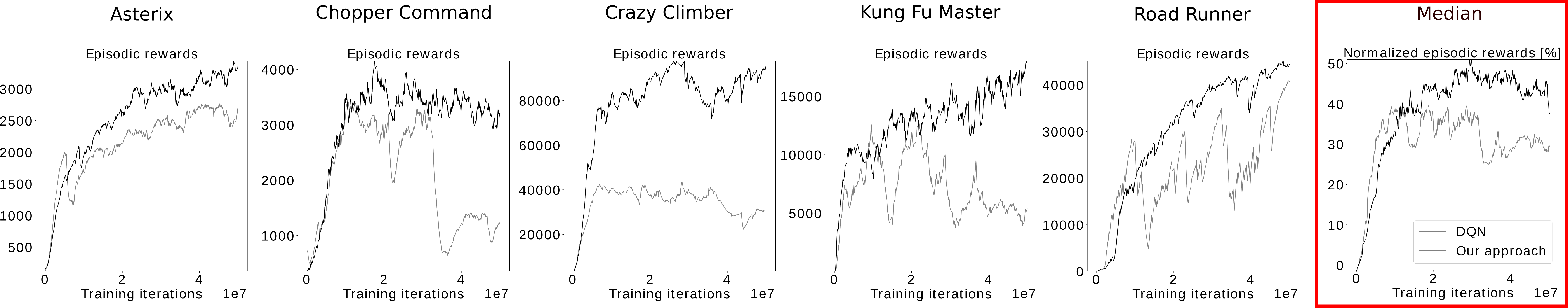}}
\caption{Individual and median game score. The plots compare the game score of our approach (black) to DQN (gray) as a function of training iterations. The left plots report game scores on individual Atari games. The rightmost plot shows the median game score across 
20  Atari games. The median plot reports human-normalized scores in order to average over games. Our approach is significantly better than DQN on each individual game and on average.}
\label{fig:game_scores_evolution}
\end{center}
\end{figure}

In order to quantify game play performance, we store agent networks every $10^5$ steps during learning and conduct an off-line evaluation by averaging over $100$ episodes, each of which comprises at most $4,500$ steps (but terminates earlier in case of a terminal event). In evaluation mode, agents follow an epsilon-greedy strategy with epsilon = $0.05$ \cite{Mnih2015}. The results of this evaluation are depicted in Figure~\ref{fig:game_scores_evolution} for five individual Atari games and for the median score over all 20 Atari games (smoothed with an exponential window of $10$). The median is taken by normalizing raw scores with respect to human and random scores according to $\text{score}_\text{norm} = \frac{\text{score}_{\text{raw}} - \text{score}_{\text{random}}}{\text{score}_{\text{human}} - \text{score}_{\text{random}}}$
following standard literature \cite{Mnih2015,Hessel2018}. 

Our approach significantly outperforms the baseline (vanilla DQN, \cite{Mnih2015}) on each individual game depicted and on average over all 20 games in the course of training. 

Additionally, we report normalized game scores obtained by the best-performing agent throughout the entire learning process (see Table~\ref{tab:max_game_scores}). Notably, our best-performing agent outperforms the DQN baseline on 14 out of 20 individual games tested.

\begin{table}[t]
\caption{Normalized episodic rewards in 20 Atari games.}
\label{tab:max_game_scores}
\centering
\begin{tabular}{ |c||c|c| }
\hline
Game & DQN & Our approach \\ 
\hline
\hline
Amidar & \textbf{22.3\%} & 19.5\%  \\
Assault & \textbf{34.1\%} & 33.3\%  \\
Asterix & 37.4\% & \textbf{46.4\%}  \\
Battle Zone & 20.2\% & \textbf{83.7\%}  \\
Berzerk & 11.6\% & \textbf{12.3\%}  \\
Chopper Command & 54.9\% & \textbf{86.2\%}  \\
Crazy Climber & 234.3\% & \textbf{378.7\%}  \\
Kangaroo & \textbf{242.6\%} & 224.5\%  \\
Krull & \textbf{713.2\%} & 663.9\%  \\
Kung Fu Master & 91.1\% & \textbf{115.8\%}  \\
Ms Pacman & 28.2\% & \textbf{32.4\%}  \\
Qbert & 72.8\% & \textbf{93.8\%}  \\
Road Runner & 582.2\% & \textbf{615.5\%}  \\
Robotank & \textbf{416.8\%} & 122.6\%  \\
Seaquest & 11.3\% & \textbf{13.6\%}  \\
Space Invaders & 50.0\% & \textbf{50.5\%}  \\
Star Gunner & \textbf{511.8\%} & 185.7\%  \\
Time Pilot & -13.7\% & \textbf{5.8\%}  \\
Tutankham & 161.1\% & \textbf{167.8\%}  \\
Up'n Down & 66.5\% & \textbf{72.7\%}  \\
\hline
\hline
Median & 60.7\% & \textbf{85.0\%}  \\
\hline
\end{tabular}
\end{table}

\subsection{Sample Efficiency}
\label{sample_efficiency}

\begin{figure}[ht]
\begin{center}
\centerline{\includegraphics[width=0.8\columnwidth]{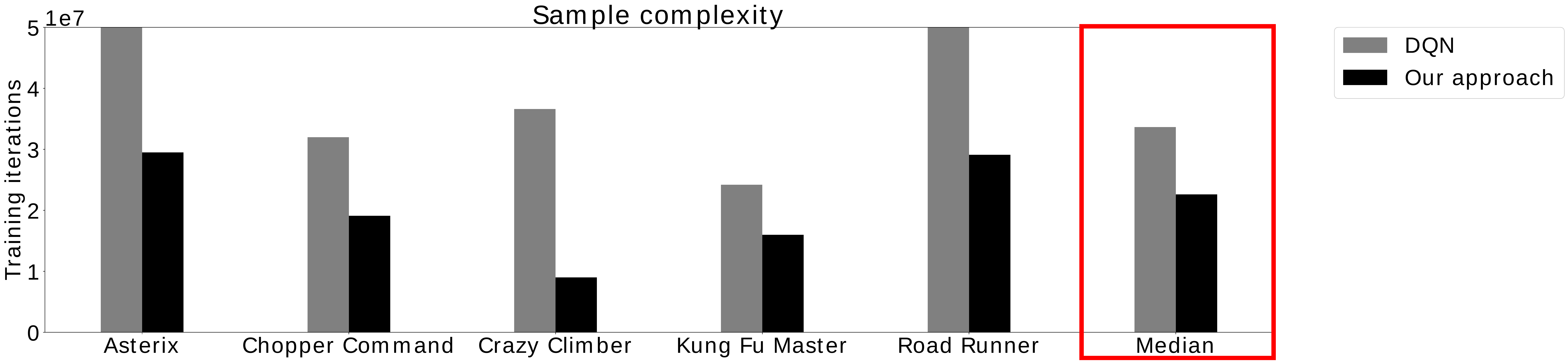}}
\caption{Individual and median sample complexity. Sample complexity is measured in terms of the number of environment interactions at which best DQN performance is attained first. Five individual Atari games are depicted on the left comparing DQN (gray) to our approach (black). The rightmost plot shows the median sample complexity over all 20 games. Our approach is more sample-efficient on each individual game depicted and on average.}
\label{fig:sample_complexity}
\end{center}
\end{figure}

In order to demonstrate sample efficiency, we identify the number of time steps at which maximum DQN performance is attained first. For this purpose, we smooth the episodic reward obtained via off-line evaluation (Figure~\ref{fig:game_scores_evolution}) with an exponential window of size $100$. The results are shown in Figure~\ref{fig:sample_complexity} for five individual games and for the median over all 20 games. Our approach achieves better sample complexity than the DQN baseline on each of the five games depicted and on average.

\section{Conclusion}
\label{conclusion}

We introduced a new optimization objective for value-based deep reinforcement learning by extending conventional deep Q-networks with a model-learning component resulting in a transcoder network.  Model prediction errors of this model are used as a regularizer for the value learning objective. We hypothesized that combining a predictive model with model-free RL leads to a richer training signal in the course of learning leading to both improved sample efficiency and overall performance.

We empirically confirmed our hypothesis by comparing to vanilla DQNs \cite{Mnih2015} without model-based regularization on the Arcade Learning Environment \cite{Bellemare2013}. Our approach yields significant improvements over DQN. Notably, our approach attains superior median normalized game score over DQN across 20 Atari games and also outperforms DQN on 14 out of 20 individual games when measuring success in terms of the best-performing agent. We also confirm improved sample efficiency by counting the number of steps at which maximum DQN performance is attained first, both when considering individual games as well as when averaging over all 20 games.

\bibliography{transcoder}
\bibliographystyle{unsrt}

\end{document}